\documentclass[10pt,journal,compsoc]{IEEEtran}
\usepackage{xcolor,soul,framed} 

\colorlet{shadecolor}{yellow}
\usepackage[pdftex]{graphicx}
\graphicspath{{../pdf/}{../jpeg/}}
\DeclareGraphicsExtensions{.pdf,.jpeg,.png}
\usepackage[cmex10]{amsmath}
\usepackage{mdwmath}
\usepackage{mdwtab}
\usepackage{eqparbox}
\newcolumntype{P}[1]{>{\centering\arraybackslash}p{#1}}
\usepackage{url}
\usepackage[caption=false]{subfig}
\usepackage{array, booktabs}
\usepackage{amsmath,amssymb,amsfonts}
\usepackage{mathtools}
\usepackage{multirow}
\usepackage{makecell}
\usepackage{algorithmic}
\usepackage[]{algorithm2e}
\usepackage{pifont}
\usepackage{verbatim}
\usepackage{subfig}
\usepackage{tabularx}
\usepackage{textcomp}
\usepackage{float}
\usepackage{cite}
\usepackage[normalem]{ulem} 
\begin{document}

\title{LTC-SUM: Lightweight Client-driven Personalized Video Summarization Framework Using 2D CNN}

\author{

Ghulam~Mujtaba, Adeel~Malik, and Eun-Seok Ryu,~\IEEEmembership{Senior Member,~IEEE}
\IEEEcompsocitemizethanks{ \IEEEcompsocthanksitem G. Mujtaba is with C-JeS Gulliver Studios, Seoul, 10390, Republic of Korea\protect\\
E-mail: gmujtabakorai@gmail.com.
\IEEEcompsocthanksitem A. Malik is with the Department of Communication System, EURECOM, Sophia-Antipolis, 06904, France.\protect\\
E-mail: adeel\_malik91@yahoo.com.
\IEEEcompsocthanksitem E.-S. Ryu is with the Department of Computer Science Education, Sungkyunkwan University, Seoul 03063, Republic of Korea.\protect\\
E-mail: esryu@skku.edu.
}

}

\markboth{}
{Mujtaba, Malik, and Ryu: LTC-SUM: Lightweight Client-driven Personalized Video Summarization Framework Using 2D CNN}



\IEEEtitleabstractindextext{%
\begin{abstract}
This paper proposes a novel lightweight thumbnail container-based summarization (LTC-SUM) framework for full feature-length videos. This framework generates a personalized keyshot summary for concurrent users by using the computational resource of the end-user device. State-of-the-art methods that acquire and process entire video data to generate video summaries are highly computationally intensive. In this regard, the proposed LTC-SUM method uses lightweight thumbnails to handle the complex process of detecting events. This significantly reduces computational complexity and improves communication and storage efficiency by resolving computational and privacy bottlenecks in resource-constrained end-user devices. These improvements were achieved by designing a lightweight 2D CNN model to extract features from thumbnails, which helped select and retrieve only a handful of specific segments. Extensive quantitative experiments on a set of full 18 feature-length videos (approximately 32.9 h in duration) showed that the proposed method is significantly computationally efficient than state-of-the-art methods on the same end-user device configurations. Joint qualitative assessments of the results of 56 participants showed that participants gave higher ratings to the summaries generated using the proposed method. To the best of our knowledge, this is the first attempt in designing a fully client-driven personalized keyshot video summarization framework using thumbnail containers for feature-length videos. Our code and trained models are publicly available at \url{https://github.com/iamgmujtaba/LTC-SUM}.
\end{abstract}

\begin{IEEEkeywords}
client-driven, personalized media, video summarization, thumbnail containers, 2D CNN.
\end{IEEEkeywords}}

\maketitle

%
\IEEEdisplaynontitleabstractindextext

\section{Introduction}\label{sec:level1}
\IEEEPARstart{I}{n} recent years, we have witnessed an exceptional growth in multimedia content, with a significant proportion of multimedia content comprised of videos. This growth trend is believed to continue in the future at even higher rates mainly because of two factors: (i) a steady increase in users’ engagement with smart and computationally powerful video recording devices, and (ii) the widespread use of social media networks and video sharing platforms as a means of communication for billions of users ~\cite{cisco2020cisco}. This tremendous growth has increased the demand for technologies that enable users to quickly browse through vast and ever-growing videos and retrieve the content of their interest. The development of autonomous video summarizing techniques is one way to achieve these goals. These techniques produce a short version of a full-length video that conveys meaningful segments. Accordingly, viewers can quickly obtain an overview of the entire story without watching the full-length video. For instance, a 90-min video of a soccer match can be summarized in a few minutes, highlighting meaningful events such as free kicks and penalty shootouts.

\begin{figure}[t]
\centering
\includegraphics[width=\linewidth,keepaspectratio]{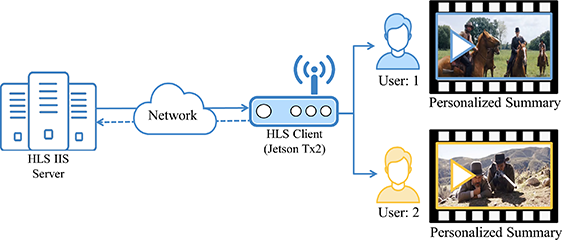}
\caption{\label{fig:conceptual} Conceptual diagram of proposed LTC-SUM framework using 2D CNN. It can generate distinct video summaries concurrently according to user preferences.}
\end{figure}

Over the last few decades, several approaches have been proposed to automate video summarization. In general, these techniques fall into two categories: keyframes ~\cite{lei2018action, thomas2018context, huang2019novel, ma2020similarity} and keyshots ~\cite{wang2012event, song2016click, zhou2018deep, fajtl2018summarizing, yuan2017video, ji2019video, apostolidis2020ac}. Keyframes are also known as static stories, representative frames, or static image summaries; in contrast, keyshots can be referred to as video skims, dynamic storyboards, or dynamic image summaries. The keyframe-based method selects a small number of image sequences from the original video, which presents an approximate visual representation. The keyshots consisted of typical continuous video segments of the full-length video that were shorter than the original video. The keyframe can be obtained from the keyshot summary in some cases ~\cite{song2016click}. In general, keyframe-based summaries are lighter in size than keyshot-based summaries. However, this gain is achieved at the cost of neglecting valuable information during the summarization process. For example, obtaining the context of the previous frame in a keyframe-based summary is challenging. In addition, it lacks the original sound. Consequently, keyshot-based video summarization methods are predominantly selected to overcome these challenges.

Keyshot-based video summarization methods are used to produce subsets for short-form videos (i.e., user-generated TikToks and news) or long-form videos (i.e., feature-length films and soccer matches). Generally, the lengths of short- and long-form videos are lesser and more than 10 min, respectively ~\cite{google_video}. As the playback duration of short-form videos is already very concise, it is impractical and may be ineffective in generating keyshot-based subsets for such videos. Moreover, the playback duration for long-form videos can exceed 90 min, specifically for movies or sports videos. Keyshot-based summary methods are more practical and effective for such video categories to provide a quick glimpse to users.

\textit{\textbf{Computational bottleneck: }}Each video contains a variety of information such as character appearance, motion, interactions between objects, events, and scenes. Considering a 1-h long-form video at 25 frames per second (FPS), it comprises thousands of frames. Existing approaches require extensive computational resources to properly process the entire video data (i.e., the frames) ~\cite{song2016click, zhou2018deep, fajtl2018summarizing, apostolidis2020ac}. If the video has an overly high definition, the demand for computing resources will increase. Deep learning-based methods also require segmented processing of long-form videos, further increasing the number of processing steps and computational complexity ~\cite{lei2018action}. Thus, these types of approaches may not be suitable for resource-constrained devices, as the device must process all frames, which increases the overall computational time. Considering that computational resources are limited, lightweight keyshot-based summarization methods for the long-form are lacking.

\textit{\textbf{Privacy bottleneck: }}Video summarization is a daunting task because of its subjectivity. This is because every user has different preferences, even for similar video content. The personalized video summarization method provides precise solutions to this problem ~\cite{babaguchi2004personalized}. The algorithm is aimed at generating customized content for every user according to their interests. However, personalized video summaries with optimal lengths for new long-form videos (e.g., sports matches) are not immediately available. With current approaches ~\cite{thomas2018context, lei2018action}, generating personalized summaries in real time requires enormous computational resources to process user preference data and video content. Centralized dedicated servers can provide real-time, personalized video summaries. However, server-based personalized solutions would require the server to have access to users’ preference data along with video content, which could lead to users’ privacy concerns.

In the context of computational and privacy bottlenecks, we propose a client-driven approach to create personalized keyshot video summaries on resource-constrained devices.

\subsection{Our Contribution}
This paper proposes a novel client-driven framework called LTC-SUM that uses lightweight thumbnail containers in the summarization process. It handles the complex process of detecting personalized events (such as penalty shoot-out in soccer videos) from lightweight thumbnails. This makes the proposed approach computationally efficient because the entire video is not processed. In addition, the technique is efficient in terms of communication (between the server and the client) and storage requirements, as the entire video does not need to be transmitted over the network and stored. Contrary to previous keyshot-based methods ~\cite{wang2012event, song2016click, zhou2018deep, fajtl2018summarizing, yuan2017video, ji2019video, apostolidis2020ac}, this study was aimed at generating subsets for long-form videos such as movies and documentaries. The proposed approach is a fully client-driven application that can generate distinct video summaries autonomously for concurrent users according to their interests (see Figure \ref{fig:conceptual} for an example). The main contributions of this study are summarized as follows:

\begin{itemize}
\item A novel thumbnail-based client-driven framework is proposed to generate keyshot video summaries according to user preference. The proposed LTC-SUM framework aims to resolve the bottlenecks of computation resources and user privacy.
\item To the best of our knowledge, this is the first study to develop a complete client-driven technique for creating personalized video summaries using thumbnail containers.
\item A lightweight two-dimensional convolutional neural network (2D CNN) model was designed to identify personalized events from thumbnails.
\item Quantitative and qualitative evaluations were conducted on full long-form eighteen videos (approximately 32.9 h in duration). Extensive quantitative experiments showed that the proposed method is more computationally efficient than the SoA baseline methods for the same client device configurations (Section \ref{sec:level4.3}). The qualitative evaluations were conducted with the collaboration of 56 participants (Section \ref{sec:level4.4}).
\end{itemize}


The remainder of this paper is organized as follows. Section \ref{sec:level2} provides a summary of related work. Section \ref{sec:level3} discusses the proposed lightweight client-driven personalized video summarization approach. A detailed implementation of the video summarization framework along with the experimental results and discussion is presented in Section \ref{sec:level4}. Finally, Section \ref{sec:level5} summarizes the paper and provides concluding remarks.

\begin{table*} [t]
\centering
\setlength{\tabcolsep}{3pt}
\caption{\label{tab:related_work} Comparison of proposed framework with exiting well-known video summarization methods.}

\begin{tabular}{|l| P{40pt} | P{40pt} | P{40pt} | P{45pt} | P{40pt} | P{40pt} | P{45pt} | P{45pt} |}
\hline
\multicolumn{1}{|c|}{\multirow{2}{*}{Methods}} & \multicolumn{4}{c|}{Content Analysis} & \multicolumn{2}{c|}{Video length} & \multicolumn{2}{c|}{Summary Type} \\ \cline{2-9} 
\multicolumn{1}{|c|}{}                         & Video  & Frames  & Segments  & Thumbnails & Short-form               & Long-form              & Keyframe              & Keyshot              \\ 
\Xhline{3\arrayrulewidth}
Song,   Yale, et al. ~\cite{song2016click}                           &        & \ding{51}       &           &    & \ding{51}                &                &   \ding{51}              & \ding{51}               \\ \hline
Yuan,   Yitian, et al. ~\cite{yuan2017video}                         &        & \ding{51}       &           &    & \ding{51}                &                &                 & \ding{51}               \\ \hline
Lei,   Jie, et al. ~\cite{lei2018action}                             &        &         & \ding{51}         &    &                  & \ding{51}              & \ding{51}               &                 \\ \hline
Thomas,   Sinnu Susan, et al. ~\cite{thomas2018context}                  & \ding{51}      &         &           &    & \ding{51}                &                & \ding{51}               &                 \\ \hline
Zhou,   Kaiyang, et al. ~\cite{zhou2018deep}                        &        & \ding{51}       &           &    & \ding{51}                &                &                 & \ding{51}               \\ \hline
Fajtl,   Jiri, et al. ~\cite{fajtl2018summarizing}                         &        & \ding{51}       &           &    & \ding{51}                &                &                 & \ding{51}               \\ \hline
Huang,   Cheng, et al. ~\cite{huang2019novel}                        & \ding{51}      &         &           &    & \ding{51}                &                & \ding{51}               &                 \\ \hline
Ji,   Zhong, et al. ~\cite{ji2019video}                           &        & \ding{51}       &           &    & \ding{51}                &                &                 & \ding{51}               \\ \hline
Ma,   Mingyang, et al. ~\cite{ma2020similarity}                         &        & \ding{51}       &           &    & \ding{51}                &                & \ding{51}               &                 \\ \hline
Apostolidis,   Evlampios, et al. ~\cite{apostolidis2020ac}               &        & \ding{51}       &           &    & \ding{51}                &                &                 & \ding{51}               \\ 
\Xhline{3\arrayrulewidth}
\textbf{Proposed}                              &        &         &           & \ding{51}  &                  & \ding{51}              &                 & \ding{51}               \\ 
\Xhline{3\arrayrulewidth}
\end{tabular}
\end{table*}

\subsection{Notations and Definitions}
The following definitions are used throughout this paper: 
\begin{itemize}
\item \textbf{Segment ($Seg$):} A video is a combination of sequences of distinct segments $Seg$ (or chunks), where the duration of each segment is a few seconds.
\item \textbf{Frame:} The video consists of a sequence of individual moving images, each of which is called a frame.
\item \textbf{Event/Action:} Event/action corresponds to certain types of activity, such as penalty shoot-out in a soccer match or horse-riding in western movies.
\item \textbf{Thumbnail container ($ThuCon$):} Thumbnail container $ThuCon$ is a collection of thumbnails extracted from the video. The sequence of all $ThuCon$ covers the entire video length.
\item \textbf{Thumbnail ($Thum$):} Thumbnail $Thum$ is obtained from the video frame. A single $ThuCon$ has 25 $Thum$, which is used in the video player to instantaneously preview the video\footnote{The number of $Thum$ in a $ThuCon$ can be varied. However, the number of $Thum$s was fixed to 25 in this study, based on our study on web-based YouTube player.}.
\end{itemize}

\section{Related Research}\label{sec:level2}
This section briefly reviews existing works on keyshot-based and personalized video summarization methods. It also reviews existing action recognition methods, which are important for identifying personalized events from thumbnails in our proposed method.

\subsection{Keyshot Based Summarization}\label{sec:level2.1}

The main idea of video summarization is to generate a short version of the original video. Video summarization techniques are divided into two categories i) keyframes ~\cite{lei2018action, thomas2018context, huang2019novel, ma2020similarity} and ii) keyshots ~\cite{wang2012event, song2016click, zhou2018deep, fajtl2018summarizing, yuan2017video, ji2019video, apostolidis2020ac}. The keyframe summarization methods generate a quick glimpse of the video as a set of images, however a significant amount of valuable information is omitted. Keyshot summarization methods attempt to overcome this challenge and provide more informative summaries in video form. Wang et al. ~\cite{wang2012event} proposed a web-based event-driven video summarization method using tag localization and keyshot mining. Initially, the tags associated with each video are localized and included in its shots, and the relevance of the shots for the event query is estimated. A set of keyshots is then classified from the shots by performing near duplicate keyframe detection. Song et al. ~\cite{song2016click} used an aesthetic measurement to detect the segmentation point changes from a video. They used the K-nearest neighbor algorithm for clustering to remove redundant frames. A sequential decision-making method for a video summarization task was proposed in ~\cite{zhou2018deep}. The deep summarization network (DSN) was designed to obtain the probability and selection of every frame in the video. Fajtl et al. ~\cite{fajtl2018summarizing} proposed a summarization method using a soft self-attention mechanism with two fully connected layers, with a sequence-to-sequence network. The network is used to process the CNN features of video frames and compute the frame-level importance scores. Later, this information was utilized in the relevant segment selection process from the video. A deep side semantic embedding (DSSE) model that leverages queries as a side information method is proposed in ~\cite{yuan2017video}. The DSSE architecture consists of two subnetworks, each with a unimodal autoencoder. One DSSE autoencoder encoded the video frames as input, and the other encoded the side information of the textual information associated with the video. The keyshot summary is generated by minimizing the distance between the selected video frame and side semantic information in the latent subspace. The importance of the sequence of video frames is measured using the proposed supervised-based encoder-decoder network ~\cite{ji2019video}. This information is used to generate a series of keyshots containing humans as output. The encoder uses a bidirectional long short-term memory (LSTM) network to encode contextual information between the input video frames. The decoder uses two attention-based LSTM networks. The encoder-decoder model is used to convert the frame-level importance score into a shot-level score in the summarization process. Recently, another summarization approach was proposed using a generative adversarial network (GAN) ~\cite{apostolidis2020ac}. An embedded actor-critic with a GAN model is designed to select the most important frames from the video. Subsequently, the selected frames were combined to generate a video summary.

\subsection{Personalized Video Summarization}\label{sec:level2.2}
Although the above-mentioned keyshot summarization techniques are valuable, they miss an influential element in the summarization process; that is user preferences. The personalized summarization techniques utilize preference-based events, shots, and features to create personalized summaries that correspond to users' interests. Wei et al. ~\cite{wei2007video} proposed a personalization method that adapted video content based on both client devices' resource constraints and user-provided keywords. In ~\cite{varini2017personalized}, long-form first-person tourist videos and user preferences were analyzed as input, and a subset of the video was returned as output. For each video, shot boundaries were detected using clustering and personalized saliency calculated by comparing the video segment to the user profile preferences and visual attention score. Ying et al. ~\cite{li2006techniques} used short- and long-term audio-visual temporal features to detect substories from movies. The length of the generated summary was adjusted according to user preference. The technique proposed in ~\cite{ellouze2010s} allowed users to select content, type of shots, and summary length in the personalized movie summarization process. However, movie content and user preferences are equated at the feature stage instead of at the semantic level. The emotions of viewers and their attention are used to create summaries in ~\cite{peng2011editing}. The viewer's mood is identified with the help of different facial expressions such as blinking, head, and eye movements observed while the viewer is watching a video. A summarization technique that relies on user-generated comments was proposed in ~\cite{chen2017personalized}. They used real-time comments from the movie created by the audience on different timestamps. The number of comments showed the excitement of the audience, and the content of the comments provided an idea of the current scene.

\subsection{Action recognition methods from videos}\label{sec:level2.3}
In the context of the personalized summarization process, detecting shots or events from a video according to the user's preferences is a crucial and challenging task. Thus, the detection of relevant events from thumbnails is essential in the proposed summarization method. Current action recognition techniques are used in the proposed method to complete this phase. In this context, some prominent SoA-CNN techniques were reviewed. CNN models have surpassed conventional approaches in recent studies ~\cite{tran2015learning, le2011learning, simonyan2014two, feichtenhofer2017spatiotemporal}. This is because they are more reliable and generalizable for extracting holistic features compared than are handcrafted. For this, variants and extensions of 2D CNNs and three-dimensional CNNs (3D CNNs) are applied to pictures. 2D CNNs perform only spatial operations on a single image. However, 3D CNNs can perform spatial and temporal operations while maintaining temporal dependencies between input video frames ~\cite{tran2015learning}. In ~\cite{le2011learning}, researchers used a 3D CNN with a support vector machine (SVM) and an independent subspace analysis (CNN-ISA) to identify human actions from the video. Similarly, another variant of a CNN network C3D was used to extract later-fed video features to SVM to identify the action ~\cite{tran2015learning}. Unlike previous methods, another CNN-based SoA action recognition method uses two types of streams, namely, the spatial and temporal streams ~\cite{simonyan2014two, feichtenhofer2017spatiotemporal}. The video decomposes into spatial (RGB representation) and temporal (optical flow representation) components. Subsequently, video frames were fed into two separate 3D CNNs.

The autonomous video summarization process for long-form videos in real-time on a client device is still an open problem. This is mainly because the current summarization techniques use segments ~\cite{lei2018action}, or entire video/frames ~\cite{ma2020similarity, yuan2017video, ji2019video, thomas2018context, huang2019novel, apostolidis2020ac, zhou2018deep, fajtl2018summarizing} data in a process that requires enormous computational resources, as shown in Table \ref{tab:related_work}. However, most modern end-user devices have low computational resources, and processing the entire video or frames takes a significantly long time to generate a summary on the client device. This is not a feasible real-time requirement. In addition, video summarization methods that require semantic information in the process may require mining and pre-processing steps ~\cite{yuan2017video} to obtain useful information, and information may not be publicly available for all videos, such as scripts and subtitles ~\cite{lei2018action}. This further increases the demand for computational resources and processing time. Thus, in this proposed work, lightweight thumbnail containers are used in the summarization process, which makes the computation process, communication, and storage efficient to create a real-time summary on the user end. Consequently, our proposed approach aims to resolve the computational and privacy bottlenecks of the personalized video summarization technique.

\begin{figure}[!b]
\centering
\includegraphics[width=\linewidth,keepaspectratio]{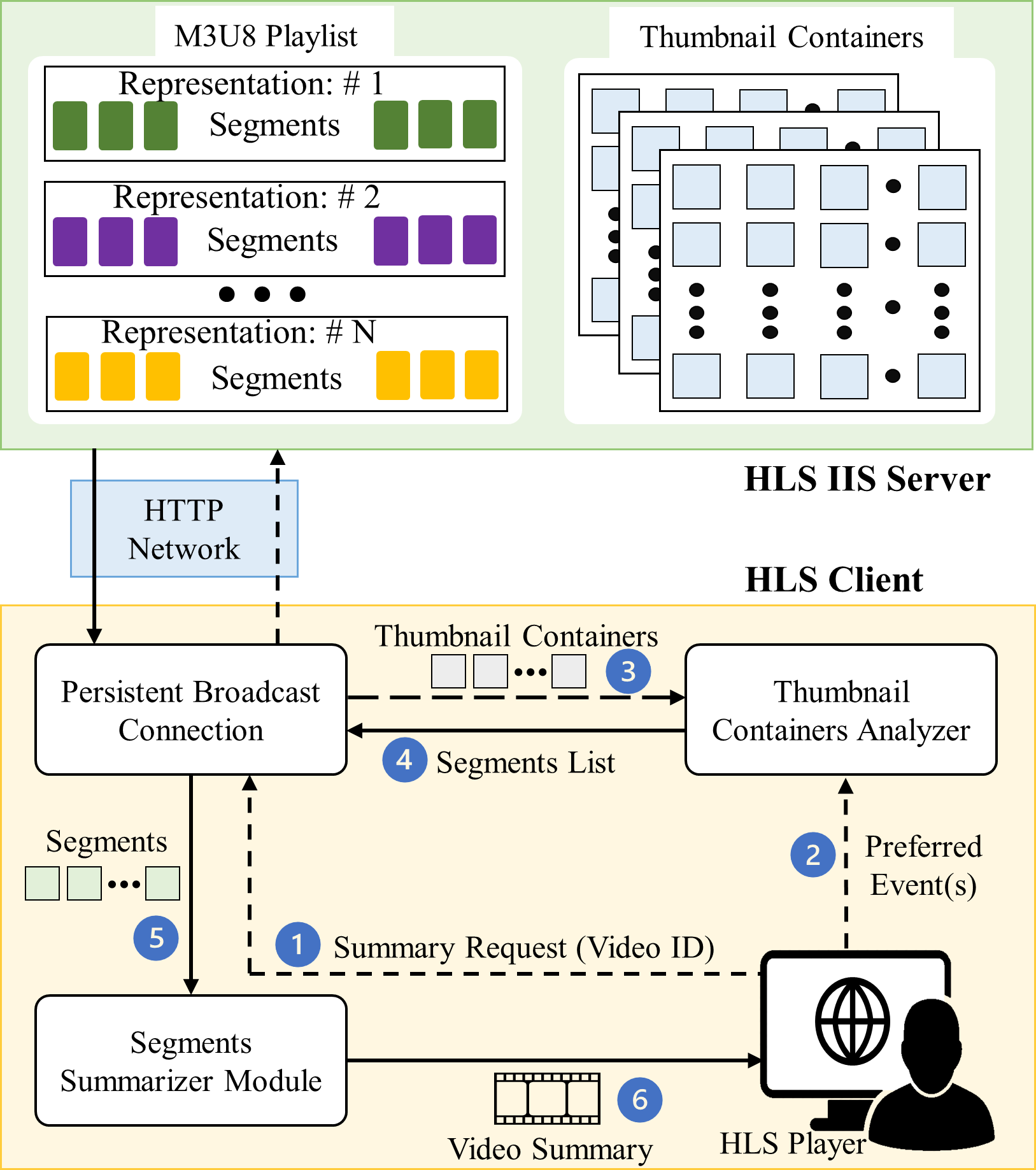}
\caption{\label{fig:system_architecture} High-level system architecture of the proposed lightweight video summarization framework.}
\end{figure}

\section{Proposed LTC-SUM Video Summarization Framework}\label{sec:level3}
A normal video is a combination of continuous moving frames at a rate of 25 fps. As described in Section \ref{sec:level2.1}, the well-known techniques that process the entire video (i.e., all frames) to generate summaries are not computationally efficient. Intuitively, for any given unit of time (i.e., one second), there exists a significant amount of redundancy in frames ~\cite{pan2018recurrent}. Thus, generating a summary by processing all frames is inefficient because unnecessarily redundant frames will also be processed, thus wasting a significant portion of the limited computational resources. Considering that computational resources are limited and expensive, it would be desirable to avoid processing frames that have a high correlation (i.e., redundant frames). In this context, a novel thumbnail-based approach is proposed to generate personalized video summaries with the aim of reducing the waste of computational resources and reducing computation time. The use of lightweight thumbnails instead of frames enables us to generate summaries within the limit of acceptable computation time for end-user devices such as the Nvidia Jetson TX2\footnote{Note that the approaches that require the processing of the entire video can also be deployed on the end-user device, but this will lead to a significantly higher computation time. This is discussed in detail in Section \ref{sec:level4.3}}.

Figure \ref{fig:system_architecture} illustrates the high-level system architecture of the proposed lightweight video summarization framework. The system comprises two main parts: the \emph{HLS IIS server} and \emph{HLS client}. In the following, we explain the configuration and role of each component of the proposed framework.

\subsection{HLS IIS Server}
\label{sec:level3.1}
The first component of the system architecture was the HLS server. The HLS server was configured locally on Microsoft Windows 10 Internet Information Services (IIS). This configuration allows multiple heterogeneous devices to concurrently download $ThuCon$ and $Seg$ for a given video from the HLS IIS server. The IIS supports a wide range of network protocols such as HTTP, HTTPS, and FTPS (refer to ~\cite{o2019iis} for the complete list). Initially, by using FFmpeg ~\cite{ffmpeg}, the entire video is encoded as the H.264/AAC MPEG-2 transport stream (.ts) segments. The MPEG-2 transport stream is suitable for transmission when there is a potential corruption or loss of data packets ~\cite{hopkins1994digital}. Each $Seg$ consists of approximately a playback portion of 10 seconds of the video, with a continuous timestamp. The text-based playlist file (M3U8) contains a list of $Seg$s according to their playback order. Each bitrate playlist contains URLs pointing to the $Seg$ files.

\begin{figure}[t]
\includegraphics[width=\linewidth,keepaspectratio]{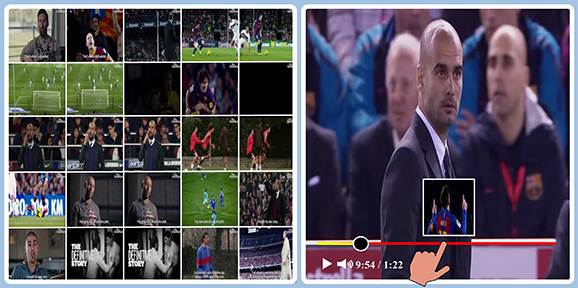}
\caption{\label{fig:thumbnail_video}Orientation of thumbnails on a single thumbnail container image (left), and the thumbnail usage for instant preview in the client web-based YouTube video player (right).}
\end{figure}

In addition to the $Seg$s, the HLS IIS server also contains $ThuCon$s, which are extracted from the corresponding video using FFmpeg ~\cite{ffmpeg}. Each $ThuCon$ has 25 $Thum$s, where a single $Thum$ of a video corresponds to the first frame of each second of the video. Thus, each $Thum$ represents one second of the video, and a single $ThuCon$ represents 25 seconds of the video. The sequence of all the $ThuCon$ covers the entire video length. Based on our study on YouTube web-based player, in this work, the size of each $Thum$ was fixed at $160\times90$ ($width\times height$) pixels and $ThuCon$ to $800\times450$ ($width\times height$) pixels. Figure \ref{fig:thumbnail_video} illustrates an example of a $ThuCon$ of a documentary received in the client web-based YouTube video player (left), and a $Thum$ previewing a particular duration (right)\footnote{The documentary was Take The Ball, Pass The Ball, and it can be obtained from URL \url{https://www.youtube.com/watch?v=VfKls9Eo1ZI}.}. Next, the configuration and the role of the HLS client are discussed.

\subsection{HLS Client}
\label{sec:level3.2}
The purpose of the HLS client is to process video-related information obtained from the HLS IIS server using the end-user computational resources and to locally generate a personalized summary of the corresponding video. For this purpose, a Nvidia Jetson TX2, which has an embedded AI computing device, is configured as an end-user computational resource. It is a GPU-based board with a Nvidia Pascal 256 CUDA core architecture along with a 64-bit hex-core ARMv8 CPU; stacked with a memory of 8 GB, and 59.7 GB/s 128-bit interface of memory data transfer capacity ~\cite{amert2017gpu}. The Jetpack 4.3 SDK is used to automate the basic installations on Nvidia Jetson TX2, which includes board support packages and libraries, especially for deep learning and computer vision. The Nvidia Jetson TX2 supports several energy profiles and the max-n profile used in the proposed approach. The HLS client consists of four major components: the persistent HTTP connection to download $ThuCon$s and personalized $Seg$s from the HLS IIS server; the deep learning-based action recognition model to recognize personalized events from thumbnails; the summarizer module to aggregate the different timestamp segments; and the web-based HLS video player for the user interface to generate a personalized summary.

\subsubsection{HTTP Persistent Connection}
\label{sec:level3.2.1}
During the generation of personalized video summary, the client initiates several requests to obtain $ThuCon$s and $Seg$ of the corresponding video from the HLS IIS server. For this purpose, a cost-effective HTTP 2.0 persistent connection was used to download $ThuCon$s and $Seg$ from the HLS IIS server. This connection enabled the exchange of numerous requests, and it returned data simultaneously in a single TCP connection. An open connection is faster for frequent data exchanges, as it remains open for HTTP requests and responses rather than closing after a single exchange. The performance of the persistent connection adaptive streaming was evaluated in ~\cite{mueller2013dynamic}. Using a persistent connection has several advantages; for example, the overall CPU usage and round trips are reduced because of fewer new connections and TLS handshakes ~\cite{zurawski2004hypertext}.


\begin{figure*}[t]
\centering
\includegraphics[width=\linewidth,keepaspectratio]{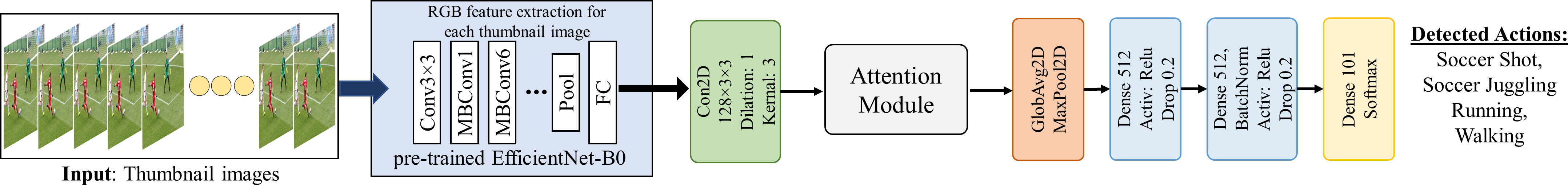}
\caption{\label{fig:deep_model} Proposed lightweight action recognition model used to analyze personalize events from thumbnails in the video summarization process. RGB thumbnails are forward propagated through a 2D CNN model to extract features from the fully connected layer.}
\end{figure*}

\begin{figure}[ht]
\centering
\includegraphics[width=\linewidth,keepaspectratio]{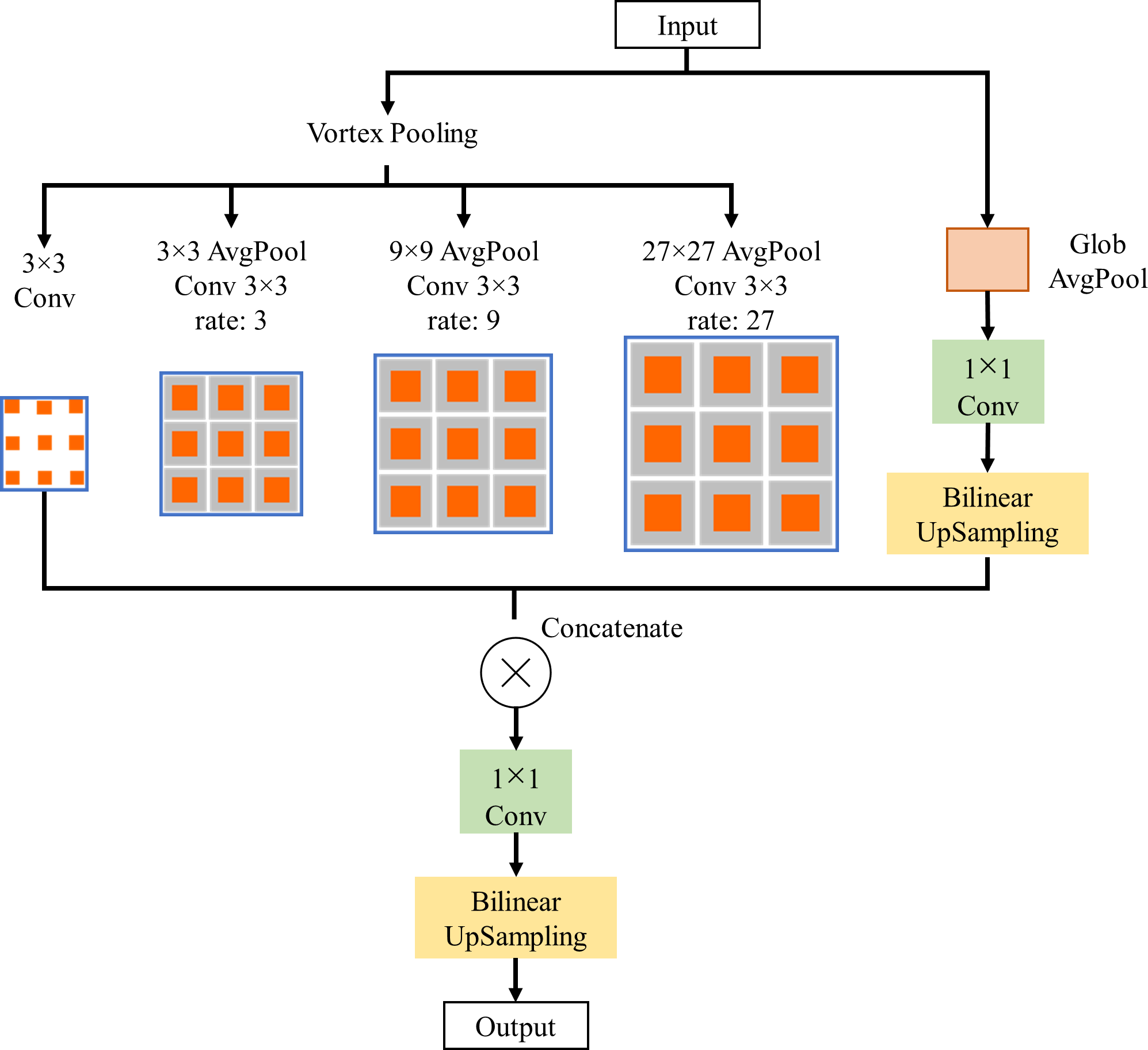}
\caption{\label{fig:vortex_poo} Architecture of vortex pooling used as an attention module in the proposed 2D CNN model.}
\end{figure}

\subsubsection{Thumbnail containers analyzer}
\label{sec:level3.2.2}
The main task of the thumbnail container analyzer is to detect the preferred events from $Thum$s. Then, based on the selected $Thum$s, generate a list of personalized $Seg$ and use them to produce a personalized summary. For this purpose, a lightweight 2D CNN model was designed to detect personalized events from each thumbnail. To detect the personalized thumbnail based on the preferred events of the user from each $Thum$ with high accuracy, the CNN model must be trained using thousands of images, which requires high processing GPU power. In this context, transfer learning ~\cite{sharif2014cnn} is useful, in which a pre-trained model is used for other purposes. This method was applied to train the EfficientNet-B0 ~\cite{tan2019efficientnet}, which was trained on a large-scale ImageNet ~\cite{deng2009imagenet} dataset to extract the frame-level spatial features of each thumbnail. Compared with other ConvNets, EfficientNet outperforms state-of-the-art architectures on ImageNet and has fewer parameters and FLOPS ~\cite{tan2019efficientnet}. This makes EfficientNet ~\cite{tan2019efficientnet} a suitable candidate for detecting personalized events from lightweight $Thum$s. The backbone of the proposed network was based on EfficientNet-B0 ~\cite{tan2019efficientnet}. Figure \ref{fig:deep_model} shows the proposed action recognition model used to process all the $Thum$s and detect personalized events. An attention module derived from vortex pooling was used to improve the performance of the proposed network ~\cite{xie2018vortex}. The attention module is more effective by using multibranch convolution with different dilation rates to aggregate contextual information. Different dilation rates can effectively improve the receptive field, consequently acquiring multilevel contextual information. The architecture of the vortex pooling used as an attention module is depicted in Figure \ref{fig:vortex_poo}.

The proposed 2D CNN model was trained on the UCF101 dataset, which is a well-known action recognition dataset ~\cite{soomro2012ucf101}. It consists of $13320$ videos taken from YouTube, which are divided into $101$ action categories ~\cite{soomro2012ucf101}. In the proposed approach, data augmentation is applied ~\cite{krizhevsky2012imagenet} to reduce overfitting; this method has been proven to be very effective. To train the model using the UCF101 dataset, each video was subsampled down to 40 frames. Before being provided as input to the network, all images were preprocessed by first cropping the center region, and then resizing them to 244$\times$244 pixels. A shear transformation was also performed at an angle of 20°, horizontal and vertical shift of 0.2, random rotation of 10°, and random horizontal flipping of images.

The dataset is split into two subsets: training and testing -- as suggested in ~\cite{soomro2012ucf101}. The model is trained using a variant of stochastic gradient descent (SGD) with a momentum of 0.9, and a learning rate of 0.01 -- using the default weight decay value (SGDW) ~\cite{loshchilov2019decoupled}. In the experiments, an early stopping mechanism was applied in the training process with a patience of ten epochs. The Keras toolbox was used for deep feature extraction, and a GeForce RTX 2080 Ti GPU was used for implementation. The training data were fed in mini batches with a size of 32 and a learning rate of 0.001 for cost minimization, and there were one thousand iterations for learning the sequence patterns in the data. The action recognition accuracy analysis of the model is presented in Section \ref{sec:level4.2}. In the following section, the third component of the HLS client is described, which is the segments summarizer module.

\subsubsection{Segments Summarizer Module}
\label{sec:level3.2.3}
The purpose of the summarizer on the client is to aggregate all the downloaded personalized $Seg$ into a single continuous video stream using FFmpeg ~\cite{ffmpeg}. The module is scalable; however, currently, it only supports continuous stream playback in the proposed approach. Note that there are no restrictions for fixing the length of the generated summary in the proposed architecture. However, as the client downloads all the personalized $Seg$, a module can be integrated into the system, which manages the summary length according to user the preference. The web-based HLS video player is explained in the following section.

\subsubsection{Web-based HLS video player}
\label{sec:level3.2.4}
\begin{figure}[t]
\includegraphics[width=\linewidth,keepaspectratio]{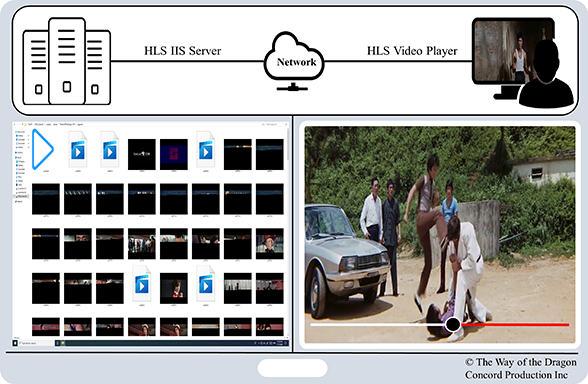}
\caption{\label{fig:video_player} HLS IIS server containing segments and thumbnail containers (left side), web-based HLS video player interface displaying generated summary (right side).}
\end{figure}

The HLS video player provides functionalities for the user to choose the video title and personalized event(s) according to their preference and then play the generated summary. The interface was designed using an open-source HTML5 HLS video player ~\cite{hlsjs}. Figure \ref{fig:video_player} shows HLS IIS server containing $Seg$s and $ThuCon$ (left side), web-based HLS video player interface displaying generated summary (right side). It supports VoD sessions, and media content (e.g., segments and playlists) can be assessed in the VoD session on the client side. The list of segments in their playback order is stored in a text-based M3U8 playlist file. The player can use the M3U8 playlist to determine the available bitrates and locations of the $Seg$s. The data delivery is entirely client-driven, which means that the video player can determine when to request each segment from the playlist file in the playback order or with a specific timestamp. In addition, it can shift between different video bitrates during playback.

\section{Experimental Results and Discussion}
\label{sec:level4}
In this section, we present an extensive experimental investigation of the proposed approach. First, the experimental setup of hardware specifications used in experiments are described. Then, the complete flow of the proposed video summary generation is explained from the user perspective. Then, the accuracy of the proposed action recognition 2D CNN model is presented and compared with those of other well-known approaches. Finally, the performance of the proposed LTC-SUM method is compared with baseline video summarization methods along with discussion.

\subsection{Experimental setup}\label{sec:level4.1}

\subsubsection{Hardware Configuration}\label{sec:level4.1.1}
In the experimental evaluation, the HLS clients and the HLS server were locally configured. Two different types of hardware configurations were used for the HLS client. The high computational resources (HCR) device was run on an open-source Ubuntu 18.04 LTS operating system with dual quad-core 2.10 GHz Xeon processors, GeForce RTX 2080 Ti, and 62 GB RAM. The low computational resources (LCR) device was a Nvidia Jetson TX2. Three distinct experimental setups were configured for the HLS client: (i) the proposed LTC-SUM approach was configured on the LCR and (ii) on HCR devices, and (iii) all the baseline approaches were configured only on the HCR device. The HLS IIS server was configured on Windows 10 in all experiments. All hardware devices were locally connected to the Sungkyunkwan University network. Table \ref{tab:hardware_specs} lists the specifications of each hardware device used in the experiments. The entire video summarization process using the proposed approach is explained in the following subsection.

\begin{table}[ht]
\centering
\caption{\label{tab:hardware_specs} SPECIFICATIONS OF HARDWARE DEVICES.}
\begin{tabular}{|P{28pt}|P{98pt}|P{57pt}|P{20pt}|}
\hline
Device & CPU & GPU & RAM \\ 
\Xhline{3\arrayrulewidth}
HCR Client & Quad-core 2.10 GHz Xeon & GeForce RTX 2080 Ti & 62 GB \\ \hline
LCR Client & HMP Dual Denver 2/2MB L2 + Quad ARM A57/2MB L2 & Nvidia Pascal 256 CUDA cores & 8 GB\\ \hline
HLS IIS Server & Intel Core i7-8700K & GeForce GTX 1080 & 32 GB \\ \hline
\end{tabular}
\end{table}


\begin{table*}[t]
\centering
\caption{\label{tab:movie_tile} List of video titles and their details used for analysis in the proposed approach.}
\begin{tabular}{|c|l| P{35pt} | P{35pt} |c| P{30pt} |c| c |c|}
\hline
S/N & \multicolumn{1}{c|}{Title} & Genre & IMDB & Duration & FPS & \# Frames & \# $ThuCon$ & \# $Thum$
\\ 
\Xhline{3\arrayrulewidth}
1  & 89 (2017)                                 & Sport   & 7.8 & 1 h 31 min & 25 & 135300  & 217 & 5412  \\ \hline
2  & Bobby (2016)                              & Sport   & 7.1 & 1 h 37 min & 25 & 1420204 & 225 & 5608  \\ \hline
3  & Bruce Lee The Man And   The Legend (1973) & Action  & 6.7 & 1 h 25 min & 24 & 123658  & 207 & 5152  \\ \hline
4  & Django (1966)                             & Western & 7.3 & 1 h 22 min & 24 & 131749  & 220 & 5495  \\ \hline
5  & Django Unchained   (2012)                 & Western & 8.4 & 2 h 45 min & 24 & 237909  & 397 & 9922  \\ \hline
6  & Goal! The Dream   Begins (2005)           & Sport   & 6.7 & 1 h 58 min & 23 & 169932  & 284 & 7087  \\ \hline
7  & Little Big Man (1970)                     & Western & 7.6 & 2 h 19 min & 24 & 2005509 & 335 & 8362  \\ \hline
8  & M.S. Dhoni The Untold   Story (2016)      & Sport   & 7.7 & 3 h 4 min  & 24 & 265933  & 444 & 11080 \\ \hline
9  & Oklahoma! (1955)                          & Western & 7   & 2 h 25 min & 24 & 201422  & 337 & 8401  \\ \hline
10 & Shanghai Noon (2000)                      & Western & 6.5 & 1 h 50 min & 23 & 158648  & 265 & 6617  \\ \hline
11 & Snake In The Eagle's   Shadow (1978)      & Action  & 7.4 & 1 h 30 min & 24 & 140473  & 235 & 5858  \\ \hline
12 & Take The Ball, Pass   The Ball (2018)     & Sport   & 8.2 & 1 h 49 min & 25 & 163972  & 263 & 6558  \\ \hline
13 & The Indian Fighter   (1955)               & Western & 6.4 & 1 h 28 min & 23 & 127126  & 213 & 5302  \\ \hline
14 & The Legend Of Drunken   Master (1994)     & Action  & 7.6 & 1 h 42 min & 24 & 147454  & 247 & 6150  \\ \hline
15 & The Rider (2017)                          & Western & 7.4 & 1 h 44 min & 24 & 148404  & 236 & 5886  \\ \hline
16 & The Train Robbers   (1973)                & Western & 6.5 & 1 h 32 min & 23 & 132192  & 221 & 5514  \\ \hline
17 & The Way Of The Dragon   (1972)            & Action  & 7.3 & 1 h 30 min & 24 & 142712  & 239 & 5953  \\ \hline
18 & Vengeance   Valley (1951)                 & Western & 5.9 & 1 h 23 min & 30 & 147448  & 197 & 4919  \\ \hline
\end{tabular}
\end{table*}

\subsubsection{Proposed Thumbnail-based Summarization Process}\label{sec:level4.1.2}
This section provides the complete flow of the proposed video summary generation process from a user perspective. This flow is described based on a set of 18 video titles used for the experiments. A complete description of the set of videos is provided in Table \ref{tab:movie_tile}. The genres of the cinematographic movies and documentaries analyzed were Western, sport, and action\footnote{The eighteen video titles arbitrarily chosen consisted of three genres (Western, sport, and action) for the experimental evaluation. However, the proposed approach is not limited to these titles and can be used for arbitrary video titles and genres.}. Since a movie/documentary may consist of more than one genre, the most dominant genre is considered (i.e., western, sport, or action). Initially, the user selects a video title from the list of available video titles using the web interface. In the experiments, the user could select a video title from among 18 video titles with different playtimes and each with the a frame size of $640\times480$ pixels.

\begin{figure}[ht]
\centering
\includegraphics[width=\linewidth,keepaspectratio]{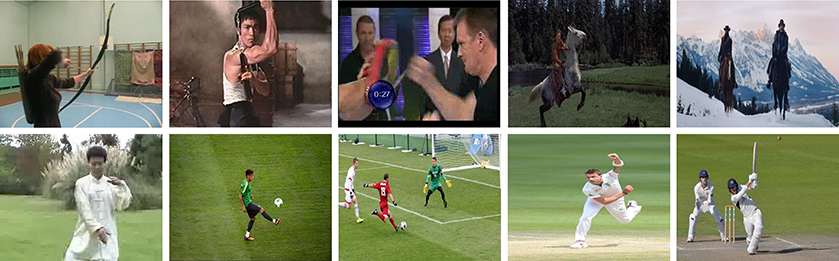}
\caption{\label{fig:user_actions}First six images display sample events for action and western genre videos: archery, nunchuck, punch, horse-race, and horse-riding. The last four images display selected events for sports genre movies: soccer-juggling, soccer-shot, cricket-bowling, and cricket-shot.}
\end{figure}

Depending on the video genre selected by the user, they were asked to choose the recommended event(s) from the list of events corresponding to the selected video genre. In the experiment, based on the set of videos described in Table \ref{tab:movie_tile}, ten distinct event(s) could be selected from the UCF101 action categories list. These events are archery, cricket-bowling, cricket-shot, horse-race, horse-riding, nunchucks, punch, soccer-juggling, soccer-shot, and tai-chi. These events were selected and categorized based on the genre of the video title. Figure \ref{fig:user_actions} shows sample images of the selected events for analyzing the video.

\begin{figure*}[t]
\centering
\includegraphics[width=\linewidth,keepaspectratio]{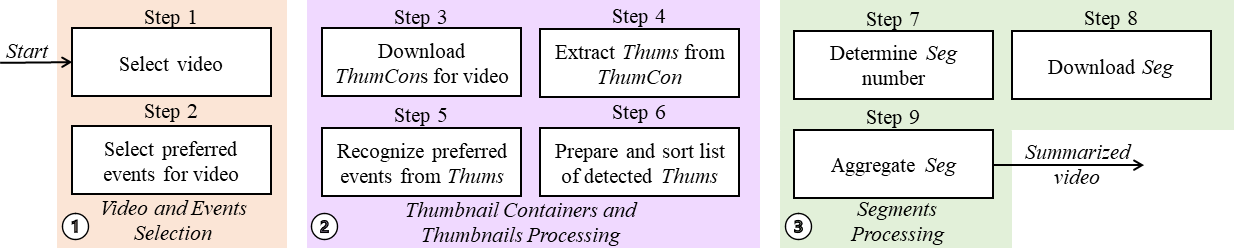}
\caption{\label{fig:summarization_module} Schematic overview of the proposed LTC-SUM summarization process.}
\end{figure*}

Once the user selects the preferred event(s), the HLS client downloads all the $ThuCon$s of the corresponding video from the HLS IIS server. Note that the downloaded $ThuCon$s cover the entire length of the video, and a very low bitrate is required to transmit all the $ThuCon$s from the server to the client. This is because $ThuCon$ tends to be lightweight in terms of size and small in number compared with the frames of the same video (refer to Table \ref{tab:movie_tile} for quick comparisons). After obtaining all the $ThuCon$s, the system extracts $Thum$s from $ThuCon$s, and the pre-trained 2D CNN model proceeds by analyzing all of them based on the preferred event(s) of the user. All $Thum$s relevant to the preferred event(s) are listed. Based on the shortlisted $Thum$s, the system generates a text-based list of detected $Thum$s in chronological order according to the $Thum$ number. The list provides temporal information about the personalized $Seg$s that need to be used to generate a personalized summary of the requested video. The text-based list of detected $Thum$s was prepared separately whenever a new process started for each video title.

The system determines the $Seg$ number from the text-based list based on the detected personalized $Thum$s, and requests to download $Seg$s with different timestamps from the HLS IIS server. If a $Seg$ takes too long to download, an alternate bitrate can be selected. Once all $Seg$s are received, the system aggregates them into one continuous video stream using FFmpeg ~\cite{ffmpeg}, in which a user can watch using the web-based HLS video player interface. The described flow of the proposed thumbnail-based summarization process is illustrated in Figure \ref{fig:summarization_module}.

\subsubsection{Baseline Approaches}\label{sec:level4.1.3}
In this subsection, the baseline approaches are described for comparison with the proposed thumbnail-based method. As explained in Section \ref{sec:level2.1}, well-known video summarization approaches process every frame of the corresponding video to generate a summary. Thus, some of the prominent SoA techniques were adopted as the baseline approaches in this study. To generate the summary using baseline methods, all videos in Table \ref{tab:movie_tile} were stored locally. However, the proposed LTC-SUM approach does not require the videos to be stored locally which brings an additional gain in storage efficiency. The baseline approaches were as follows:

\begin{itemize}
\item \textbf{HECATE ~\cite{song2016click}.} It analyzes the aesthetic features from all temporarily extracted frames of the corresponding video. This method only supports fixed summary lengths, and a five-minute subset is generated for each video.
\item \textbf{DR-DSN ~\cite{zhou2018deep}.} It is trained based on the SumMe dataset ~\cite{gygli2014creating} using the default parameters. Initially, this method extracts frames from the corresponding video and then analyzes the extracted frames to generate a video summary. Using DR-DSN with default parameters, the summarization duration generated for all corresponding videos was 22 seconds.
\item \textbf{VASNet ~\cite{fajtl2018summarizing}.} Similar to DR-DSN ~\cite{zhou2018deep}, it is trained based on the SumMe dataset ~\cite{gygli2014creating} using default parameters. First, it extracts frames from the corresponding video and then analyzes extracted frames to generate a subset. For every corresponding video, a 24-second subset is generated using the default parameters of VASNet.
\item \textbf{AC-SUM-GAN ~\cite{apostolidis2020ac}.} Similar to the aforementioned baseline methods ~\cite{zhou2018deep, fajtl2018summarizing}, it first extracts frames from the corresponding video and then analyzes the extracted frames to generate a subset. It is trained based on the SumMe dataset ~\cite{gygli2014creating}, and it generates at 19-second summary for every corresponding video with the default configuration.
\item \textbf{FB-SUM.} This is the frame-based summarization (FB-SUM) baseline method that analyzes every frame of the corresponding video during the process. Initially, the video frames were extracted using FFmpeg ~\cite{ffmpeg} for each video. The rest of the summation process followed the same steps as the proposed LTC-SUM method.
\end{itemize}

As highlighted before, there is a redundancy in frames while processing the entire video; thus, processing all frames of a video is not computationally efficient as it increases the processing time and wastes a significant portion of the limited computational resources. To validate the effectiveness, the computation time of the baseline approaches were compared with the proposed LTC-SUM approach in the following experiments.

\subsection{Experimental evaluation of action recognition datasets}\label{sec:level4.2}
In this subsection, the accuracy of the proposed action recognition 2D CNN model is evaluated using the benchmark action recognition dataset \emph{UCF101} ~\cite{soomro2012ucf101}. To the best of our knowledge, the best thumbnail-based approach was proposed in ~\cite{mujtaba2020client}, therefore, we compared our results with ~\cite{mujtaba2020client}. The proposed model reported the highest validation accuracy of 77.81\% in 36 epochs with 55.74 million flops. The proposed method achieved an increase of 4.06\% in the validation accuracy, increasing from 73.75\% ~\cite{mujtaba2020client} to 77.81\%, and the training accuracy increased from 91.41\% ~\cite{mujtaba2020client} to 96.06\%. The work in ~\cite{mujtaba2020client} used InceptionV3, which has 24M parameters; however, EfficientNet-B0 was used as the backbone network in the proposed LTC-SUM method, which has 6.9M parameters. Comparisons with the other methods are summarized in Table \ref{tab:methods_comparisons}.

\begin{table} [ht]
\centering
\setlength{\tabcolsep}{3pt}
\caption{\label{tab:methods_comparisons} COMPARISON OF AVERAGE RECOGNITION SCORE OF THE ACTION RECOGNITION PROPOSED METHOD WITH OTHER METHODS.}

\begin{tabular}{ l P{110pt} }

\hline
Methods & UCF101 

\\
\Xhline{3\arrayrulewidth}
Karpathy, Andrej, et al. 2014. \cite{karpathy2014large}& 65.4\% \\ \hline
Murthy, OV Ramana, et al. 2015 \cite{murthy2015ordered}& 72.8\% \\ \hline
Mujtaba, et al. 2020 \cite{mujtaba2020client} & 73.75\%\\ \hline
Shu, Yu, et al. 2018 \cite{shu2018odn} & 76.07\%\\ \hline
Mujtaba, et al. 2022 \cite{mujtabasigmap2022} & 76.25\%\\ \hline
Liu, An-An, et al. 2016 \cite{liu2016hierarchical} & 76.3\% \\ \hline
\Xhline{3\arrayrulewidth}
Proposed Method & \textbf{77.81\%} 
\\ \Xhline{3\arrayrulewidth}

\end{tabular}
\end{table}

\begin{table*} [t]
\centering
\setlength{\tabcolsep}{3pt}
\caption{\label{tab:jarvis_all_baseline} THE TOTAL COMPUTATION TIME REQUIRED IN MINUTES TO GENERATE A SUMMARY USING BASELINE METHODS ON AN HCR DEVICE, AND THE PROPOSED LTC-SUM METHOD ON HCR AND LCR DEVICES.}
\begin{tabular}{|c| P{60pt} | P{60pt} | P{65pt} |c| P{65pt} | P{65pt} | P{65pt} |}
\hline
\multirow{2}{*}{S/N} & HECATE ~\cite{song2016click} & DR-DSN ~\cite{zhou2018deep} & VASNet ~\cite{fajtl2018summarizing} & AC-SUM- GAN ~\cite{apostolidis2020ac} & FB-SUM & \multicolumn{2}{c|}{Proposed LTC-SUM} \\ \cline{2-8}
& \multicolumn{5}{c|}{HCR}                                                           & LCR           & HCR           \\

\Xhline{3\arrayrulewidth}
1  & 20.85 & 4.92 & 5.09 & 4.81 & 70.41  & 7.64  & \textbf{1.98} \\ \hline
2  & 22.02 & 5.15 & 4.77 & 4.87 & 67.23  & 8.16  & \textbf{1.91} \\ \hline
3  & 28.51 & 4.10 & 4.31 & 4.43 & 60.91  & 5.56  & \textbf{1.98} \\ \hline
4  & 21.05 & 5.13 & 5.32 & 4.74 & 63.67  & 8.25  & \textbf{2.10} \\ \hline
5  & 54.41 & 8.58 & 8.36 & 8.69 & 132.32 & 14.64 & \textbf{3.58} \\ \hline
6  & 28.56 & 6.32 & 6.18 & 5.92 & 80.60  & 10.03 & \textbf{2.43} \\ \hline
7  & 51.91 & 7.32 & 7.03 & 7.04 & 102.16 & 12.07 & \textbf{3.02} \\ \hline
8  & 85.23 & 9.53 & 9.93 & 9.52 & 123.29 & 15.87 & \textbf{4.22} \\ \hline
9  & 53.49 & 7.74 & 7.91 & 7.98 & 118.60 & 12.23 & \textbf{2.93} \\ \hline
10 & 32.03 & 5.51 & 5.57 & 5.64 & 82.65  & 9.85  & \textbf{2.36} \\ \hline
11 & 25.50 & 5.24 & 4.91 & 4.86 & 73.14  & 6.49  & \textbf{2.23} \\ \hline
12 & 28.29 & 6.01 & 5.88 & 6.16 & 131.83 & 9.87  & \textbf{2.31} \\ \hline
13 & 18.71 & 5.21 & 4.75 & 4.64 & 65.36  & 7.90  & \textbf{1.95} \\ \hline
14 & 31.33 & 5.23 & 5.23 & 4.98 & 77.21  & 6.78  & \textbf{2.27} \\ \hline
15 & 17.20 & 5.57 & 5.09 & 4.97 & 73.45  & 9.08  & \textbf{2.32} \\ \hline
16 & 23.11 & 4.78 & 4.77 & 4.81 & 60.82  & 7.93  & \textbf{2.06} \\ \hline
17 & 29.00 & 5.39 & 5.29 & 5.20 & 78.08  & 6.67  & \textbf{2.25} \\ \hline
18 & 24.59 & 5.55 & 5.57 & 5.08 & 74.36  & 7.64  & \textbf{1.91} \\ \hline

\end{tabular}
\end{table*}

\subsection{Performance evaluation of the proposed LTC-SUM method}\label{sec:level4.3}
This subsection compares the performance of the proposed LTC-SUM video summarization method with the baseline schemes described in Section \ref{sec:level4.1.3}. The performance evaluation experiments used ten different events to analyze the proposed LTC-SUM and baseline approaches. The list of events is described in Section \ref{sec:level4.1.2}. All the detected $Thum$s for the proposed LTC-SUM method and frames for the FB-SUM baseline method are included in the summarization process for which the detection accuracy was higher than 95\% for western, 65\% for action, 80\% for cricket sports, and 90\% for soccer sports videos\footnote{The threshold value directly impacts the duration of the generated summary. If a low threshold is selected, then a lengthy summary will be generated.}. The threshold of each video genre was selected to maintain the length of the summaries. The default parameters were used for the remaining baseline methods.

The computation time (in minutes) required to generate a video summary using the baseline and the proposed approaches were compared in the first experiment, where all approaches were configured on the HCR device (refer to Table \ref{tab:hardware_specs} for detailed specifications of the device). The steps involved in calculating the computation time are (i) frame extraction from the video (FB-SUM baseline) and $Thum$s extraction from $ThuCon$s (proposed); (ii) event(s) recognition using the lightweight trained 2D CNN model from frames (FB-SUM baseline) and $Thum$s (proposed); (iii) determining and downloading $Seg$; (iv) and finally, aggregate $Seg$ into a single continuous video stream. Meanwhile, default configurations and steps are used for HECATE ~\cite{song2016click}, DR-DSN ~\cite{zhou2018deep}, VASNet ~\cite{fajtl2018summarizing}, and AC-SUM- GAN ~\cite{apostolidis2020ac} baseline approaches to generate summaries. Compared with the number of frames, the number of lightweight thumbnail images was significantly smaller (Table \ref{tab:movie_tile}). Thus, the overall computation time of the proposed LTC-SUM method is significantly lower than that of the FB-SUM baseline approach. Table \ref{tab:jarvis_all_baseline} shows the computation time in minutes required to generate a summary using baseline methods, on the HCR device. Extracting frames from the video and using all the extracted frames to generate a summary are the key factors in increasing the overall computation time while using the baseline methods.

Because this study focused on generating summaries resource-constrained on client end devices, in the next experiment, the proposed LTC-SUM method is configured on the LCR device (i.e., the Nvidia Jetson TX2). Table \ref{tab:jarvis_frame_based_baseline} lists the computation time required in minutes on every step to generate a summary using the FB-SUM baseline method on the HCR device and proposed LTC-SUM method on the HCR and LCR devices.

Table \ref{tab:summary_length_thumbnail} depicts the duration of summaries generated automatically to detected frames/$Thum$s for the corresponding video using FB-SUM and LTC-SUM methods on the LCR device. From Table \ref{tab:jarvis_frame_based_baseline}, it can be observed that the computation time for FB-SUM is significantly higher than that for LTC-SUM. In addition, it can also be observed from Table \ref{tab:jarvis_all_baseline} that even when the proposed technique is implemented on the LCR device, the computation time is still significantly shorter than that of FB-SUM and HECATE ~\cite{song2016click} baseline approaches implemented on the HCR device.

\begin{table*} [t]
\centering
\setlength{\tabcolsep}{3pt}
\caption{\label{tab:jarvis_frame_based_baseline} COMPUTATION TIME REQUIRED IN EVERY STEP TO GENERATE THE SUMMARY USING THE FB-SUM METHOD ON THE HCR DEVICE AND THE PROPOSED LTC-SUM METHOD ON HCR AND LCR DEVICES.}

\begin{tabular}{|c|c| P{25pt} | P{25pt} |c| P{25pt} | P{25pt} |c| P{25pt} | P{25pt} |c| P{25pt} | P{25pt} |c| P{25pt} | P{25pt} |}
\hline
\multirow{3}{*}{S/N} & \multicolumn{3}{c|}{Frame/$Thum$   extraction}           & \multicolumn{3}{c|}{Events   recognition}                & \multicolumn{3}{c|}{Download   $Seg$s}                 & \multicolumn{3}{c|}{Aggregate  $Seg$s}                & \multicolumn{3}{c|}{Total}                    
\\ \cline{2-16} 
                     & \multirow{2}{*}{FB-SUM}      & \multicolumn{2}{c|}{LTC-SUM} & \multirow{2}{*}{FB-SUM}   & \multicolumn{2}{c|}{LTC-SUM} & \multirow{2}{*}{FB-SUM}   & \multicolumn{2}{c|}{LTC-SUM} & \multirow{2}{*}{FB-SUM}   & \multicolumn{2}{c|}{LTC-SUM} & \multirow{2}{*}{FB-SUM} & \multicolumn{2}{c|}{LTC-SUM} \\ \cline{3-4} \cline{6-7} \cline{9-10} \cline{12-13} \cline{15-16} 
                     &                              & LCR           & HCR          &                           & LCR           & HCR          &                           & LCR           & HCR         &                           & LCR           & HCR          &                         & LCR           & HCR          \\ 
\Xhline{3\arrayrulewidth}
1  & 4.98  & 0.14 & 0.11 & 65.27  & 7.33  & 1.82 & 0.13 & 0.15 & 0.04 & 0.02 & 0.02 & 0.01 & 70.41  & 7.64  & \textbf{1.98} \\ \hline
2  & 5.06  & 0.17 & 0.11 & 62.06  & 7.86  & 1.78 & 0.09 & 0.12 & 0.02 & 0.02 & 0.01 & 0.01 & 67.23  & 8.16  & \textbf{1.91} \\ \hline
3  & 4.72  & 0.12 & 0.09 & 56.09  & 5.34  & 1.86 & 0.09 & 0.09 & 0.02 & 0.01 & 0.01 & 0.01 & 60.91  & 5.56  & \textbf{1.98} \\ \hline
4  & 5.29  & 0.16 & 0.10 & 58.26  & 7.88  & 1.94 & 0.10 & 0.18 & 0.05 & 0.02 & 0.03 & 0.01 & 63.67  & 8.25  & \textbf{2.10} \\ \hline
5  & 10.32 & 0.28 & 0.18 & 121.87 & 14.24 & 3.36 & 0.11 & 0.11 & 0.03 & 0.02 & 0.02 & 0.01 & 132.32 & 14.64 & \textbf{3.58} \\ \hline
6  & 6.05  & 0.21 & 0.13 & 74.49  & 9.74  & 2.27 & 0.05 & 0.07 & 0.01 & 0.01 & 0.01 & 0.01 & 80.60  & 10.03 & \textbf{2.43} \\ \hline
7  & 7.58  & 0.25 & 0.16 & 94.41  & 11.56 & 2.79 & 0.15 & 0.22 & 0.06 & 0.02 & 0.03 & 0.02 & 102.16 & 12.07 & \textbf{3.02} \\ \hline
8  & 9.88  & 0.33 & 0.20 & 113.3  & 15.42 & 3.98 & 0.09 & 0.10 & 0.03 & 0.02 & 0.02 & 0.01 & 123.29 & 15.87 & \textbf{4.22} \\ \hline
9  & 8.69  & 0.23 & 0.15 & 109.82 & 11.94 & 2.75 & 0.07 & 0.05 & 0.02 & 0.02 & 0.01 & 0.01 & 118.60 & 12.23 & \textbf{2.93} \\ \hline
10 & 6.9   & 0.18 & 0.12 & 75.63  & 9.54  & 2.20 & 0.10 & 0.11 & 0.04 & 0.02 & 0.02 & 0.01 & 82.65  & 9.85  & \textbf{2.36} \\ \hline
11 & 5.17  & 0.17 & 0.11 & 67.72  & 6.05  & 2.05 & 0.22 & 0.24 & 0.06 & 0.03 & 0.03 & 0.01 & 73.14  & 6.49  & \textbf{2.23} \\ \hline
12 & 6.28  & 0.21 & 0.12 & 125.38 & 9.3   & 2.10 & 0.15 & 0.32 & 0.08 & 0.02 & 0.04 & 0.02 & 131.83 & 9.87  & \textbf{2.31} \\ \hline
13 & 5.76  & 0.15 & 0.10 & 59.5   & 7.67  & 1.81 & 0.08 & 0.07 & 0.03 & 0.02 & 0.01 & 0.01 & 65.36  & 7.90  & \textbf{1.95} \\ \hline
14 & 5.95  & 0.17 & 0.11 & 71.15  & 6.49  & 2.13 & 0.09 & 0.11 & 0.02 & 0.02 & 0.01 & 0.01 & 77.21  & 6.78  & \textbf{2.27} \\ \hline
15 & 6.01  & 0.16 & 0.12 & 67.33  & 8.66  & 2.09 & 0.10 & 0.24 & 0.09 & 0.02 & 0.03 & 0.01 & 73.45  & 9.08  & \textbf{2.32} \\ \hline
16 & 4.91  & 0.17 & 0.10 & 55.83  & 7.59  & 1.89 & 0.07 & 0.15 & 0.07 & 0.01 & 0.02 & 0.01 & 60.82  & 7.93  & \textbf{2.06} \\ \hline
17 & 5.96  & 0.16 & 0.12 & 72.02  & 6.32  & 2.08 & 0.08 & 0.17 & 0.04 & 0.02 & 0.02 & 0.01 & 78.08  & 6.67  & \textbf{2.25} \\ \hline
18 & 5.43  & 0.18 & 0.10 & 68.83  & 7.19  & 1.67 & 0.09 & 0.23 & 0.13 & 0.02 & 0.03 & 0.02 & 74.36  & 7.64  & \textbf{1.91} \\ \hline

\end{tabular}
\end{table*}


\begin{table} [t]
\centering
\setlength{\tabcolsep}{3pt}
\caption{\label{tab:summary_length_thumbnail}DURATION OF GENERATED VIDEO SUMMARIES IN THE SECOND EXPERIMENT FOR THE FB-SUM AND PROPOSED LTC-SUM METHODS.}

\begin{tabular}{|c|c|c|c|c|c|c|}
\hline
\multirow{2}{*}{S/N} & \multicolumn{2}{c|}{\#   Detected Images} & \multicolumn{2}{c|}{\#   $Seg$s Requested} & \multicolumn{2}{c|}{Summary Duration} \\ \cline{2-7} 
                     & FB-SUM                 & LTC-SUM                & FB-SUM            & LTC-SUM            & FB-SUM             & LTC-SUM            \\  
\Xhline{3\arrayrulewidth}
1  & 4157 & 67  & 105 & 34 & 18m30s  & \textbf{5m51s }  \\ \hline
2  & 2522 & 22  & 85  & 15 & 15m37s  & \textbf{2m34s }  \\ \hline
3  & 1248 & 37  & 96  & 25 & 17m34s  & \textbf{4m12s }  \\ \hline
4  & 4940 & 106 & 107 & 51 & 18m11s  & \textbf{8m55s }  \\ \hline
5  & 4108 & 62  & 103 & 30 & 18m14s  & \textbf{4m36s }  \\ \hline
6  & 810  & 15  & 45  & 13 & 7m22s   & \textbf{1m58s }  \\ \hline
7  & 3278 & 110 & 129 & 54 & 22m39s  & \textbf{9m49s }  \\ \hline
8  & 1518 & 48  & 100 & 24 & 16m55s  & \textbf{4m37s }  \\ \hline
9  & 2197 & 45  & 99  & 16 & 16m56s  & \textbf{2m52s }  \\ \hline
10 & 1600 & 52  & 89  & 29 & 15m14s  & \textbf{4m48s }  \\ \hline
11 & 5487 & 120 & 204 & 72 & 34m36s  & \textbf{13m1s }  \\ \hline
12 & 7317 & 150 & 131 & 66 & 22m40s  & \textbf{11m41s } \\ \hline
13 & 841  & 35  & 78  & 20 & 13m      & \textbf{3m17s }  \\ \hline
14 & 984  & 32  & 106 & 23 & 19m1s   & \textbf{4m2s }   \\ \hline
15 & 2858 & 130 & 75  & 43 & 13m12s  & \textbf{7m25s }  \\ \hline
16 & 1156 & 41  & 63  & 25 & 10m50s  & \textbf{4m29s }  \\ \hline
17 & 1986 & 80  & 103 & 47 & 18m40s  & \textbf{8m15s }  \\ \hline
18 & 5814 & 226 & 112 & 83 & 19m58s  & \textbf{13m54s } \\ \hline
\end{tabular}
\end{table}

Considering that the combined duration of all videos was 1,974 min, HECATE ~\cite{song2016click}, DR-DSN ~\cite{zhou2018deep}, VASNet ~\cite{fajtl2018summarizing}, AC-SUM-GAN ~\cite{apostolidis2020ac}, and FB-SUM baseline approaches took 595.789 min, 107.28 min, 105.96 min, 104.34 min and 1536.09 min using the computational resources of the HCR device to generate the 18 summaries for each video as shown in Table \ref{tab:jarvis_all_baseline}, respectively. Meanwhile, the proposed approach on HCR took 43.82 min to generate the 18 summaries for each video. Thus, based on the analysis of these 18 videos, computationally for the HCR device, on average, the proposed approach is 13.59, 2.45, 2.42, and 2.38 times faster than HECATE ~\cite{song2016click}, DR-DSN ~\cite{zhou2018deep}, VASNet ~\cite{fajtl2018summarizing}, and AC-SUM-GAN ~\cite{apostolidis2020ac}, respectively. The computational resources of the LCR device are very low compared to the HCR device, even when the proposed method is 3.57 HECATE ~\cite{song2016click}; and the 9.2 FB-SUM method is faster than the baseline approaches on the LCR device. In conclusion, these results show that the proposed method is computationally efficient even for an LCR device.

Note that the proposed approach is also efficient in terms of communication and storage compared to the baseline approaches. As in the baseline approaches, the complete video needs to be downloaded and stored. In the proposed approach, only the $ThuCon$s are downloaded and stored. Thus, compared with the complete video, the download time and storage requirements for $ThuCon$s are significantly less. For example, the size of the movie 89 (2017) is approximately 612 MB, while the size of the $ThuCon$s of the corresponding movie is just approximately 14 MB. In addition, DR-DSN ~\cite{zhou2018deep}, VASNet ~\cite{fajtl2018summarizing}, AC-SUM-GAN ~\cite{apostolidis2020ac}, and FB-SUM baseline approaches need to store the original video along with the extracted frames during the summarization process. By comparing number of $Thum$s with number of frames in a video, the number of frames is very large. Thus, significant local storage is needed for the baseline approaches. Therefore, in addition to achiving the computational efficiency, the proposed LTC-SUM method is also efficient in terms of storage and communication requirements for the summarization process.

From Tables \ref{tab:jarvis_all_baseline}-\ref{tab:summary_length_thumbnail}, it can be concluded that the proposed approach is significantly better than the baseline approaches in terms of low computational complexity and processing time for long-form videos. This superiority exists even when the proposed approach is configured on a significantly LCR device (Nvidia Jetson TX2). Interestingly, the duration of the summaries generated using the proposed approach was much smaller than the duration of the summaries generated using FB-SUM baseline approach (refer to Table \ref{tab:summary_length_thumbnail}). It is intuitive to ask what the impact of the significant reduction in computational time and the small duration of video summaries has on the quality of the summary. In the following section, the results of a comprehensive qualitative survey are presented to answer this question.

\subsection{Qualitative evaluation}\label{sec:level4.4}
This section presents an evaluation of the quality of the summaries generated using the proposed method by comparing it with the summaries generated using the FB-SUM baseline approach. Because this study focused primarily on personalized summaries, only the FB-SUM baseline approach was evaluated. The evaluation was based on a survey conducted with the help of $56$ participants: $44$ males and $12$ females with an age range of $15$\textendash$35$ years-- (i.e., most respondents were young). The participants were from nine different geographical locations and covered a wide range of professions; however, most respondents were researchers and faculty members.

The survey was based on 18 movies (refer to Table \ref{tab:movie_tile}), depending on the genre of the video, and a list of options for event(s) was defined. The participants could choose to generate a personalized summary. The selected options of event(s) from the UCF101 dataset for Western genre videos were (i) horse-riding, horse-racing, (ii) archery, punch, and (iii) horse-riding, horse-racing, archery, and punch. For action genre videos: (i) archery, punch, (ii) tai-chi, nunchuck, and (iii) tai-chi, nunchuck, archery, and punch. The sports genre videos were divided into two categories:-- soccer and cricket. The selected options of event(s) for soccer genre videos were (i) soccer-juggling, (ii) soccer-penalty, and (iii) soccer-juggling and soccer-penalty. For cricket genre videos: (i) cricket-bowling, (ii) cricket-shot, and (iii) cricket-bowling and cricket-shot\footnote{Note that the proposed technique is not limited to the above-mentioned list of preference options for each video. For simplicity, we adopted the event from the UCF101 dataset and defined a list of preference options for each video. Proposing a sophisticated method that can generate a list of preference options is beyond the scope of this study.}.

Each participant selected one of the movie titles and the corresponding option of event(s) from the list. For each movie title and the preferred option of event(s), two summaries were generated using the proposed LTC-SUM and FB-SUM baseline techniques. The performance of the generated video summaries was evaluated objectively using an exact rating scale. The participants were asked to rate the summary, which was considered better according to the three evaluation criteria: information coverage, visual pleasure, and general satisfaction. An anonymous questionnaire was created for the generated summaries so that the users could not determine which method (i.e., LTC-SUM or FB-SUM) was used. They were requested to watch both summaries and answer questions by ranking the results on a scale of 1--10 (1 being the worst and 10 being the best). Table \ref{tab:subjective_analysis} lists the questions and the average ratings given by the participant for each question for both approaches.

\begin{figure*}[t]
\centering
\includegraphics[width = 18.1cm]{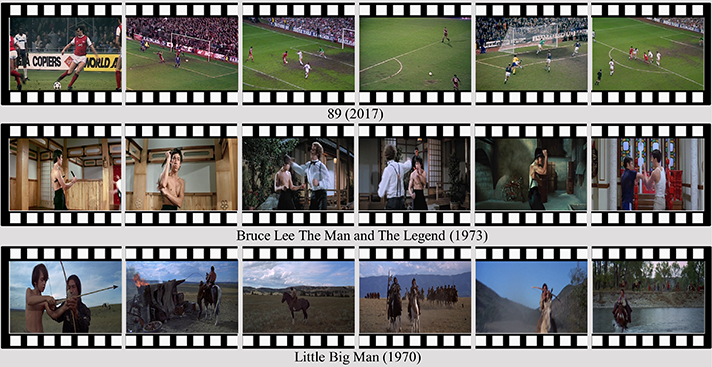}
\caption{\label{fig:summary_sample} Illustrations of the frame samples from generated video summaries.}
\end{figure*}

\begin{table} [t]
\centering
\setlength{\tabcolsep}{3pt}
\caption{\label{tab:subjective_analysis} AVERAGE RATING (1$\sim$10) OF THE FB-SUM BASELINE AND PROPOSED LTC-SUM APPROACHES.}
\begin{tabular}{|l|c|c|}
\hline
\multicolumn{1}{|c|}{Questions}                                                                                                           & Baseline  & Proposed      \\ 
\Xhline{3\arrayrulewidth}

\begin{tabular}[c]{@{}l@{}}Q1: Did the generated summary give related \\actions    (events) according to your preferences?\end{tabular} & 7.14      & \textbf{7.59} \\ \hline
Q2: Rate   generated summary.                                                                                                             & 7.16      & \textbf{7.52} \\ \hline
\begin{tabular}[c]{@{}l@{}}Q3: Is the   length appropriate for the generated\\    summary?\end{tabular}                                   & 6.45      & \textbf{7.39} \\ \hline
\begin{tabular}[c]{@{}l@{}}Q4: Compare   to both generated summaries \\which one is good rate, please.\end{tabular}                    & 6.89      & \textbf{7.32} \\ \hline
\begin{tabular}[c]{@{}l@{}}Q5: Correlations   (similarities) of the generated\\    summaries.\end{tabular}                                & \multicolumn{2}{c|}{6.89} \\ \hline
\begin{tabular}[c]{@{}l@{}}Q6: Would   you like to watch the movie after\\    watching the generated summary?\end{tabular}                & 7.09      & \textbf{7.14} \\ \hline
\end{tabular}
\end{table}

Despite the fact that the summary generated using the proposed approach was short and required less computation time as the entire analysis was based only on $Thum$s, the qualitative evaluation suggests that the proposed approach was almost the same (better in some aspects) compared with the FB-SUM baseline approach. From the results of Q1--Q2, we can say that the proposed method does not lose the personalized aspects in terms of the preferred events compared with the FB-SUM baseline. From Q3--Q4, it can be observed that the length of the summary is crucial, as most users prefer short summaries, thus leading to significantly higher average ratings for the proposed approach. In Q5, we specifically asked about the similarities among the summaries of both approaches, and the obtained results suggest that participants observed significant similarities with an average rating of 6.89. Based on this qualitative evaluation, it can be concluded that the proposed approach performs very well and receives higher average ratings compared with the FB-SUM baseline without losing significant important information (e.g., preferred events). Figure \ref{fig:summary_sample} depicts the sample frames obtained from the video summary generated using the proposed LTC-SUM method.


\subsection{Discussion and Limitations}\label{sec:level4.5}
In the previous sections, we evaluated the overall effectiveness by comparing the proposed LTC-SUM method with the baseline approaches. The proposed framework exhibited a better performance by lowering the computational complexity and computation time in the summarization process. During quantitative experiments, it was observed that there were many redundant images (frames) to determine the segment numbers using the FB-SUM baseline approach. Thus, a significant portion of the computational resources are used to process the redundant frames and determine the segment number from the detected frames. Meanwhile, the proposed LTC-SUM method needs to process fewer images (thumbnails) to determine the segment number, as shown in Table \ref{tab:summary_length_thumbnail}. This significantly reduces the computational time, the demand for computational resources, communications, and storage required to generate summaries. In addition, the proposed method can solve the computational and privacy bottlenecks on the resource-constrained end-user devices during the personalized summarization process.

During the qualitative evaluation in Section \ref{sec:level4.4}, the average ratings of the summaries generated using the proposed approach were higher than the FB-SUM baseline. One of the reasons that summaries generated using the proposed approach have higher ratings is that they have a short duration. The summary generated using the proposed LTC-SUM and FB-SUM baseline approaches have similar events. Table \ref{tab:subjective_analysis} lists the similarity ratings provided by the participants for the generated summaries. The proposed LTC-SUM approach can generate summaries according to user interests with a highly computationally efficient mechanism.

Previously, full long-form videos were segmented into small clips duration in the summarization process ~\cite{lei2018action}. This is because extensive computational resources are required to store temporal information while analyzing complete long-form videos. However, the overall computational complexity is increased by adding more processing steps to generate a summary. The overall computation, communication, and storage efficiencies are improved significantly by analyzing thumbnail containers using the proposed method. It can extract and generate summaries based on user preferences from relevant content (such as events and objects). However, it might not be effective for short-form videos that have multiple and faster scene transitions. It was observed that sometimes some frames in the generated summary are not relevant according to the preferred event(s) -- which can be mitigated by adopting the solution suggested in previous research ~\cite{mujtaba2020client}.

Currently, this paper focuses on the full long-form videos of three genres -- Western, sports, and action. The simplicity and scalability for implementing different configuration devices made it easy to adapt the proposed framework to other genres of video. In addition, it can support privacy-preserving solutions ~\cite{newell2013design} effectively by adapting efficient encryption techniques ~\cite{mujtaba2019}; it can be adapted to three screen TV solutions ~\cite{mujtaba2020gif,mujtaba2021client,ryu2011home, ryu2008towards}. The proposed method can also be adopted in ATSC 3.0 and can use over-the-top (OTT) services to provide a personalized interactive application.

\section{Concluding Remarks}\label{sec:level5} 
This paper presents a personalized lightweight client-driven LTC-SUM keyshot video summarization framework. The framework is designed for resource-constrained end-user devices to generate personalized summaries using their computational resources while resolving computational and privacy bottlenecks. Instead of using entire video data, which are computationally intensive, the lightweight thumbnail containers are used in the proposed method to generate subsets of the corresponding video. This significantly improves the computational, communication, and storage efficiencies as compared to state-of-the-art summarization approaches. For this purpose, a lightweight 2D CNN model was designed to detect personalized events from thumbnails. Extensive quantitative experiments were conducted on full 18 feature-length videos that demonstrated the superior performance of LTC-SUM compared to several state-of-the-art video summarization approaches using the same computational resources end-user devices. Qualitative results showed that the proposed method outperformed the baseline approach and received higher average ratings without losing significantly important information. It is planned to integrate the proposed method with other streaming protocols as future work.



%

\appendices


\normalem
\bibliographystyle{IEEEtran}
\bibliography{IEEEabrv,manuscript_r1}



\end{document}